\documentclass[conference]{IEEEtran}
\IEEEoverridecommandlockouts
\usepackage{cite}
\usepackage{amsmath,amssymb,amsfonts}
\usepackage{algorithmic}
\usepackage{graphicx}
\usepackage{textcomp}
\usepackage[dvipsnames,svgnames,x11names]{xcolor}
\usepackage[colorinlistoftodos,prependcaption]{todonotes}
\usepackage[inline]{enumitem}
\usepackage{hyperref}

\makeatletter 
\newcommand{\linebreakand}{%
  \end{@IEEEauthorhalign}
  \hfill\mbox{}\par
  \mbox{}\hfill\begin{@IEEEauthorhalign}
}
\makeatother 

\def\BibTeX{{\rm B\kern-.05em{\sc i\kern-.025em b}\kern-.08em
    T\kern-.1667em\lower.7ex\hbox{E}\kern-.125emX}}
\begin{document}

\title{Exploring Machine Learning Algorithms for Infection Detection Using GC-IMS Data: A Preliminary Study}

\author{
\IEEEauthorblockN{Christos Sardianos}
\IEEEauthorblockA{\textit{Dept. of Informatics and Telematics} \\
\textit{Harokopio University of Athens}\\
Athens, Greece \\
sardianos@hua.gr}
\and
\IEEEauthorblockN{Chrysostomos Symvoulidis}
\IEEEauthorblockA{\textit{Research \& Development Dept.}\\
\textit{Telesto Technologies}\\
Athens, Greece \\
csymvoulidis@telesto.gr}
\and
\IEEEauthorblockN{Matthias Schlögl}
\IEEEauthorblockA{\textit{Solgenium GmbH} \\
Linz, Austria \\
m.schloegl@solgenium.com}
\and

\linebreakand 
\IEEEauthorblockN{Iraklis Varlamis}
\IEEEauthorblockA{\textit{Dept. of Informatics and Telematics} \\
\textit{Harokopio University of Athens}\\
Athens, Greece \\
varlamis@hua.gr}
\and
\IEEEauthorblockN{Georgios Th. Papadopoulos}
\IEEEauthorblockA{\textit{Dept. of Informatics and Telematics} \\
\textit{Harokopio University of Athens}\\
Athens, Greece \\
g.th.papadopoulos@hua.gr}
}

\maketitle

\begin{abstract}
The developing field of enhanced diagnostic techniques in the diagnosis of infectious diseases, constitutes a crucial domain in modern healthcare. By utilizing Gas Chromatography-Ion Mobility Spectrometry (GC-IMS) data and incorporating machine learning algorithms into one platform, our research aims to tackle the ongoing issue of precise infection identification. Inspired by these difficulties, our goals consist of creating a strong data analytics process, enhancing machine learning (ML) models, and performing thorough validation for clinical applications. 
Our research contributes to the emerging field of advanced diagnostic technologies by integrating Gas Chromatography-Ion Mobility Spectrometry (GC-IMS) data and machine learning algorithms within a unified Laboratory Information Management System (LIMS) platform. 
Preliminary trials demonstrate encouraging levels of accuracy when employing various ML algorithms to differentiate between infected and non-infected samples. Continuing endeavors are currently concentrated on enhancing the effectiveness of the model, investigating techniques to clarify its functioning, and incorporating many types of data to further support the early detection of diseases.

\end{abstract}

\begin{IEEEkeywords}
Infection detection, GC-IMS, machine learning, data-driven healthcare
\end{IEEEkeywords}

\section{Introduction}
\subsection{Background and Motivation}
The efficient and precise diagnosis of infectious diseases is still a major obstacle in healthcare, requiring the creation of new technology and analytical frameworks. Gas Chromatography-Ion Mobility Spectrometry (GC-IMS) is a highly promising analytical method that can detect volatile organic compounds (VOCs) in complicated biological samples like breath, urine, and blood. 
The VOC profiles provide essential information that can be used as biomarkers for certain diseases, providing a non-invasive and fast method for detecting and monitoring diseases. Additionally, there is a scarcity of research in which volatile compounds, have been employed to ascertain viral infection \cite{gould2020breath}.

Although GC-IMS has the capability to detect biomarkers, there are difficulties associated with analyzing and interpreting data, especially in datasets with a large number of dimensions. Conventional analytical methods face difficulties in dealing with the intricate and unpredictable character of biomarker signatures, emphasizing the requirement for sophisticated data analytics and machine learning techniques. The objective of this work is to close this divide by creating an advanced Laboratory Information Management System (LIMS) platform that incorporates machine learning techniques, to facilitate streamlined data processing, discovery of biomarkers, and classification of diseases. These innovations have the potential to improve diagnostic accuracy, advance customized medicine, and enhance patient care outcomes. In this manuscript, we present the initial steps in this direction, as well as the early results of our experiments on using machine learning algorithms to identify infections over breath samples.

\subsection{Objectives of the Research}

The main goal of this research is to define a concrete approach for integrating sophisticated diagnostic tools with machine learning techniques in the task of infection detection and biomarker discovery from breath samples. In this direction, it aims to create a Laboratory Information Management System (LIMS) platform that is specifically designed for analyzing high-dimensional data from Gas Chromatography-Ion Mobility Spectrometry (GC-IMS) instruments. This platform is specifically developed to coordinate and manage the entire process of handling data, which includes collecting, organizing, preparing, and analyzing data. 

The first step in any data mining or machine learning process is the preprocessing of data, which can significantly improve the performance of the whole approach. In this direction,  the platform seeks to ensure data quality and consistency for subsequent analysis by implementing common pre-processing stages such as artifact detection, denoising, baseline correction, and data transformation. Data quality is the basis for prediction accuracy, so an infrastructure that automates the data quality  control process is crucial for the detection tasks that follow. To support non-expert users, this objective includes the creation of an intuitive user interface and automated functionalities for data gathering and interpretation. 
The next step is to include cutting-edge machine learning algorithms into the LIMS platform to enhance data analysis and classification. The platform aims to accurately differentiate between infected and non-infected samples by utilizing decision trees, logistic regression, partial least squares discriminant analysis (PLS-DA), random forest, and support vector machines (SVM). These methods, apart from giving a prediction on the task, they can also provide explanations about their decisions, which are very useful to the doctors. Feature selection, is also a crucial aspect of infection detection, and plays a pivotal role in identifying the most informative features within high-dimensional datasets, thereby enhancing the accuracy and interpretability of machine learning models applied to the task.

This research aims on filling important gaps in the early detection or prediction of infectious diseases and methods for public health response. The primary objective is to create a prognostic instrument that can identify diseases at an early stage, and enable well-informed clinical decision-making. This will contribute to the advancement of precision medicine projects and enhance early infection prediction and treatment outcomes.


\section{Literature Review}
The growing need for quick and precise infection identification in clinical settings has led to major developments in the field of biomarker detection in recent years. Precision medicine aims to identify, diagnose, and treat subtypes of heterogeneous diseases based on genetics and environments. Under this scope, the authors in \cite{climer2023machine} introduce a machine-learning biomarker evaluation method called Difference in Bicluster Distances (DBD) for screening real-valued data to identify biomarkers associated with sub-types of heterogeneous diseases. It utilizes clustering of data points to identify significant genes and appropriate thresholds for biomarker diagnostics. Authors in \cite{xie2022novel} proposed a novel feature selection method based on a graph neural network that considers feature dependence and the real pathway relationship of genes. The method combines a graph structure and Pearson correlation coefficient to construct a graph structure model, uses the information propagation and aggregation method of graph neural networks to characterize node information, and applies the spectral clustering method to filter redundant features. The discovery of critical biomarkers is significant for clinical diagnosis, drug research and development.

In order to examine high-dimensional data and find novel biomarkers that can distinguish between different types of illnesses, researchers have been experimenting with different machine learning and artificial intelligence methods. For example, authors in \cite{torun2022transparent} describe the development of "OmicLearn," an open-source browser-based ML tool for biomarker discovery from proteomics and omics data. It utilizes the latest advances in the Python ML ecosystem and allows easy access to ML without programming or bioinformatics skills. Data matrices from omics experiments can be uploaded to an online or locally installed web server, enabling rapid exploration of the suitability of various ML algorithms for the experimental data sets. In the same context, \cite{ding2023biomarker} proposes a deep neural network called DFMbpe that uses binary pairwise encoding to identify disease-related biomarkers by considering the interdependence of features. DFMbpe combines wide low-order feature interdependence obtained through factorization and deep high-order feature interdependence obtained through deep neural networks to achieve better performance than existing identification models. 

In the context of SARS-CoV-2 data analysis, authors in \cite{steppert2021rapid} demonstrated that ion mobility spectrometry coupled with a multicapillary column (MCC-IMS) can detect SARS-CoV-2 infection and Influenza-A infection in breath samples. The study analyzed the nasal breath of 75 patients, with 16 patients being positively diagnosed for SARS-CoV-2 infection and 14 patients for Influenza-A.


\section{Methodology}

\begin{figure*}[htbp]
\centering
\includegraphics[width=0.7\linewidth]{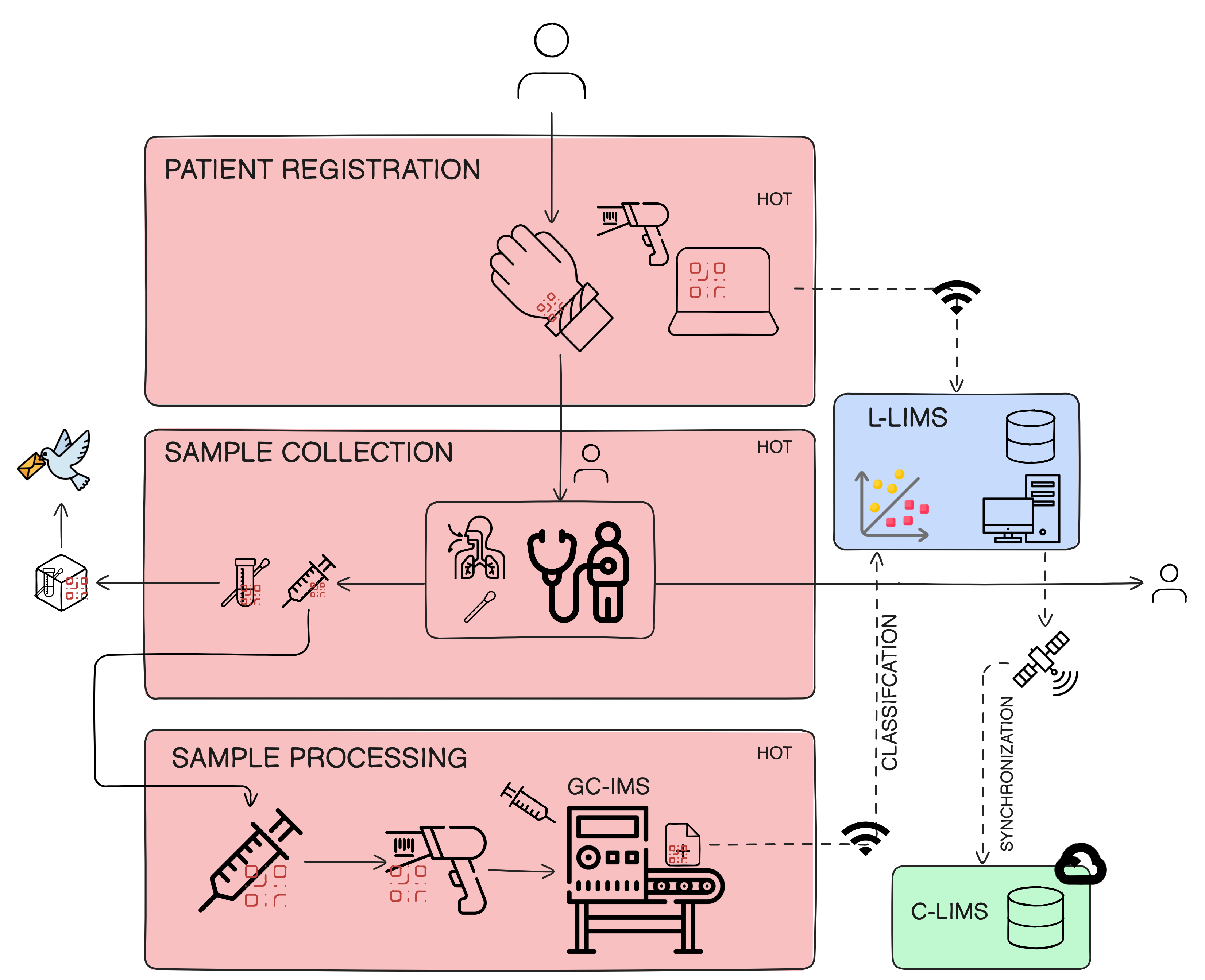} 
\caption{The proposed pipeline used in the LIMS platform for data collection, processing and analysis.}
\label{fig:lims-pipeline}
\end{figure*}


\subsection{Overview of the Platform}

The "Laboratory Information Management System (LIMS)" platform proposed in this work\footnote{The LIMS platform is part of the ONELAB European Research project, ONELAB: Orchestrating next-generation mobile modular laboratories for pandemic monitoring preparedness – EU HORIZON - grant agreement No. 101073924} (Figure \ref{fig:lims-pipeline}) automates the processes of data collection, verifying the quality of the data, and using the data to train Machine Learning (ML) models as well as utilizing them for inference. It has been divided into two main components: (i) the centralized component, whose primary goal regards the collection of data towards ML training; and (ii) the decentralized component, deployed within an Rapid Response Mobile Laboratory (RRML), where it can be used for the collection of samples from patients, as well as for the utilization of the ML models for inference which have been previously trained using the data of the centralized component. In the following subsections, a detailed analysis of the relevant functionalities related to data collection and preprocessing and the training of the ML models will be presented.

\subsection{Data Collection and Preprocessing}
To start with the data collection process, the proposed platform allows the upload of anonymized analytical data (i.e., patient measurements, clinical data, etc.). These data can be analyzed further within the platform and the functionalities it provides, while it can be used for the training of ML models which will be described in the following sections. In order to assure the quality of the uploaded data and ensure that they are interoperable and correctly structured, upon the upload of any type of data, specific quality checks are performed. 

Different types of information (e.g., measurement data, meta data, infection type, and clinic files) are collected from different sites. The data is uploaded to the centralized component of the proposed platform by the appointed local data manager after an authentication is made. In order to protect the privacy of the volunteers, the details of the patient measurements and all of the data uploaded is anonymized, meaning that it will not contain any personal information. This allows the utilization of these data for ML training without violating privacy regulations.

As mentioned above, specific quality assurance checks are performed once data are uploaded in the platform. These checks are specific to the type of data that have been uploaded every time. More specifically, when it comes to GC-IMS (Gas Chromatography-Ion Mobility Spectrometry) data, data validation checks are retrieved from the measurement file that accompanies the GC-IMS measurement and corresponding information is displayed to the local data manager in order to inform them of the quality check's outcomes. Furthermore, GC-IMS data upload can only be performed if the corresponding Clinic Report File (CRF) is valid. The CRF template contains built-in checks that all need to be successfully passed. If not, the data cannot be uploaded. When it comes to metadata quality checking, the proposed platform controls for valid data ranges (e.g., $ age < 100 $) and categories (e.g., “male”, “female”) prior to their upload. Finally, the proposed platform stores the data in Databases and/or Hierarchical Data Format (HDF) depending on their type. 

\subsection{Understanding GC-IMS data principles}
GC-IMS is particularly useful for applications in chemical analysis, environmental monitoring, food safety, and security. GC-IMS is a hybrid analytical technique used to separate and analyze complex mixtures of volatile compounds that combines the capabilities of gas chromatography (GC), that is used to separate a mixture of compounds into individual components based on their chemical properties, such as volatility and affinity for a stationary phase, and ion mobility spectrometry (IMS) that is used to further separate and identify compounds based on their size, shape, and charge. A compound is a substance composed of two or more different elements chemically bonded together in fixed proportions. In simpler terms, it is a substance formed when atoms of different elements combine through chemical reactions to create a distinct and stable chemical entity with its own unique properties. 

\begin{figure}[htbp]
\centering
\includegraphics[width=0.6\linewidth]{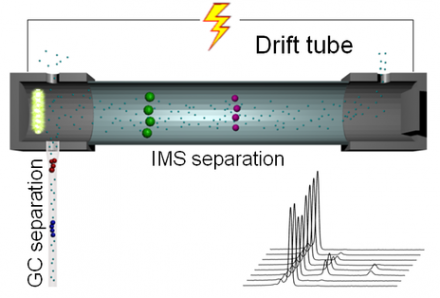} 
\caption{Schematic representation of the GC-IMS system with separation of VOCs in gas chromatography component and ion mobility. (Image source: www.odournet.com)}
\label{fig:gcims}
\end{figure}

Gas chromatography is a well-established separation technique that separates volatile compounds based on their interaction with a stationary phase inside a chromatographic column. Ion mobility spectrometry is a technique that separates ions based on their mobility through a drift tube filled with a buffer gas under the influence of an electric field (Figure \ref{fig:gcims}). 

The data generated by the GC process include:
\begin{itemize}
  \item Retention Time: The time it takes for each compound to travel through the GC column, which is unique to each compound and can be used for identification.
  \item Peak Intensity: The intensity or abundance of each compound's signal as it elutes from the GC column, which is related to its concentration in the sample.
\end{itemize}

The data generated from the IMS process include:
\begin{itemize}
  \item Drift Time: The time it takes for ions generated from the separated compounds to travel through a drift tube under the influence of an electric field. Different ions experience different drift times based on their size and charge, allowing for additional separation.
  \item Ion Mobility Spectra: The resulting spectra, often represented as plots or histograms, showing the distribution of ion drift times. Peaks in the IMS spectra correspond to different ions and can aid in compound identification.
\end{itemize}


\subsection{GC-IMS data analysis}

In GC-IMS, the eluted compounds from the GC column are introduced into the IMS drift tube as a continuous flow of vaporized molecules. An ion detector at the end of the IMS drift tube measures the arrival times of ions, which are used to create an ion mobility spectrum. The ion mobility spectrum represents the ions' mobility and provides information about the compounds present in the sample. By correlating the elution time from the GC column with the ion mobility information from the IMS, GC-IMS generates a two-dimensional separation, allowing for enhanced compound identification and characterization. 

Upon this processing, further steps of analysis can be used in order to classify and detect possible infections in the collected samples. This analysis in most cases involve the identification of specific clinical features that can indicate the presence of a disease. Towards this direction, machine learning techniques can play an important role. For this process, after collecting and pre-processing the data, feature selection techniques need to be applied to identify the most relevant features (e.g. genes, proteins etc.). This step, usually involves employing statistical tests (e.g. ANOVA) and/or correlation analysis. At this point, we need to choose the appropriate machine learning algorithms for the task of classifying the input samples as infected or non-infected. Among the algorithms that are frequently used for such classification tasks are:
\begin{enumerate}
  \item \textbf{Decision Trees} can be used to identify patterns or feature thresholds, 
  \item \textbf{Random Forest} has been used for feature selection, classification, and biomarker discovery. It's effective at handling noisy data and identifying relevant features (peaks), 
  \item \textbf{Support Vector Machines} have been employed for biomarker detection in GC-IMS data by learning to distinguish between samples with and without specific biomarkers based on their ion mobility spectra, 
  \item \textbf{Deep Belief Networks (DBNs)} that use multiple layers of stochastic, latent variables and can be used for feature extraction and classification in high-dimensional data, and 
  \item \textbf{Deep Learning model} (e.g. CNNs for image data), that can be applied for feature extraction and classification of GC-IMS data, and to learn complex patterns in ion mobility spectra.
\end{enumerate}

After choosing the appropriate model based on the evaluated performance metrics and the complexity and interpretability potentials of each model \cite{smolinska2014current, raschka2018model, mahesh2020machine} (in case of two or more techniques perform with similarly, the most interpretable one is chosen), we proceed with training the selected model on the training data (the features extracted from the GC-IMS data can be useful), and fine-tune hyper-parameters (e.g. using techniques like grid search or random search) to optimize its performance. Additionally, we also employ techniques like cross-validation (e.g., k-fold cross-validation) to assess the model's generalization ability and to prevent overfitting, applying for example regularization techniques (e.g., L1, L2 regularization), especially for high-dimensional data. 

Next, we use relevant evaluation metrics to assess the model's performance, depending on the nature of the task (binary or multi-class classification). Common metrics include accuracy, precision, recall (sensitivity), F1-score, and metrics specific to peak-based analyses (e.g. peak accuracy, ROC-AUC, or PR-AUC) are also considered. Returning back to the pre-processing step and experimenting with different pre-processing techniques, such as peak alignment or deconvolution, may improve data quality and model performance.

\section{Experimental Setup}

In the realm of healthcare and medical research, classification algorithms hold significant importance, particularly in the context of infection detection. Identifying and classifying infections accurately can have profound implications for disease prognosis, diagnosis, and treatment. Classification algorithms, such as support vector machines, random forests, and deep neural networks, are leveraged to analyze complex biological data, including genomics, proteomics, and clinical data \cite{freire2021full, teli2023application, altaf2023artificial, srivastava2023explainable}. Additionally, classification algorithms are instrumental in predicting patient outcomes and personalizing treatment plans based on individual profiles, thereby advancing precision medicine and improving healthcare outcomes. Under this scope, the GC-IMS data provided by the ONELAB consortium offers a valuable resource for the detection of infection. The primary approach for harnessing the potential of the GC-IMS data used in the scope of this research for the task of infection detection involves the application of machine learning models for classification. These models are instrumental in categorizing breath samples into two distinct groups: \textbf{infected} and \textbf{non-infected}. By leveraging these algorithms, we can exploit the inherent patterns and relationships within the input features to facilitate accurate sample classification. 
Our overarching objective is to train robust classification models using the provided GC-IMS data, ultimately yielding a predictive tool capable of efficiently determining sample infection status that could be integrated in the LIMS platform.

\subsection{GC-IMS Data Description}

\begin{figure*}[htp]
\centering
\includegraphics[width=0.8\linewidth]{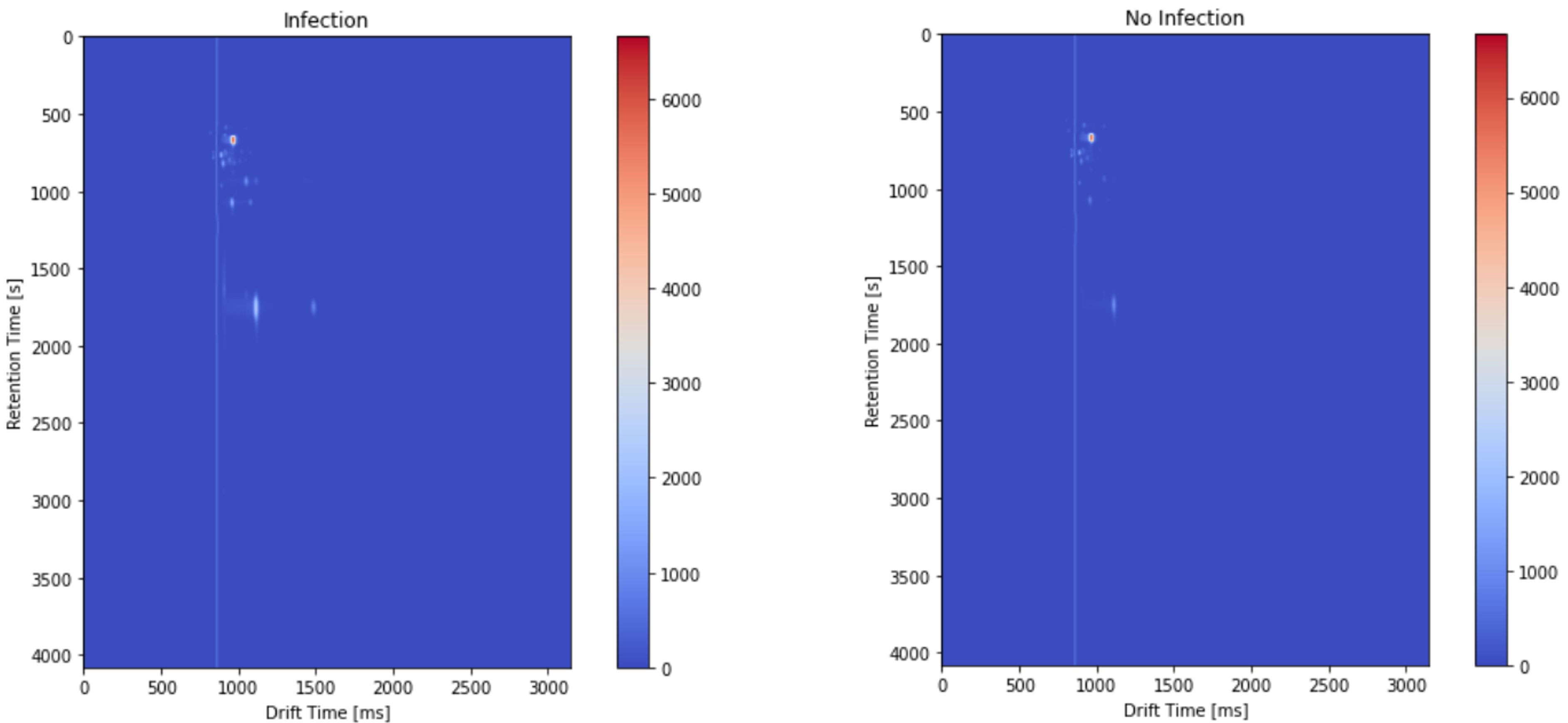} 
\caption{Visualization of one instance per class (Infection/No infection) in the GC-IMS data samples used for the classification algorithms development.}
\label{sample1_spectra}
\end{figure*}

In order to experimentally evaluate the performance of our proposed approach on identifying infections on breath samples, a set of GC-IMS data were used for developing and evaluating most of the well-known classification algorithm, as they were described before. The data sample consisted of 76 pre-processed instances of GC-IMS data with manually assigned class labels. In general, 76 instances are considered a small set for training ML algorithms, but since this is a preliminary study to examine the POC of this platform, in the full scale scenario the dataset to be used will be enhanced to cover the needs of the proper model training. Each data instance is represented by a matrix with a dimension of: 4080*3150 in which the X dimension is the Drift Time and the Y dimension is the Retention Time while the actual values of the matrix represent the intensity of the concentration of each compound in each Drift-Retention Time point. Figure \ref{sample1_spectra} illustrates the spectra of two instances, one for each class (Infection/No infection).
The axis values used for the retention time and drift time refer to the corresponding dimensions of the samples and are used only for demonstration purposes. Given the high number of features in the dataset, dimensionality reduction is crucial to reduce the computational complexity, while still preserving the relevant information for the classification. For this purpose, Principal Component Analysis (PCA) was employed in this setup for dimensionality reduction, resulting to 84-304 selected features depending on the algorithm selected from those tested in the next section. Finally, data were split into train-test (80\%-20\%) and 5-fold cross validation was performed for tuning.

\subsection{Machine Learning Models and Classification}

In order to perform classification tasks on the GC-IMS data for infection detection, we employed a variety of state-of-the-art machine learning methods. Our initial approach focused on testing the feasibility of this study, using only fundamental machine learning algorithms, some of which are self-explainable, and on a second stage we aim on extending our work examining other techniques like neural networks and more deep methods. To guarantee a thorough study of the data, a variety of approaches were included in the algorithms chosen for review. Among these techniques are: 
\begin{itemize}
  \item Decision Trees, 
  \item Logistic Regression, 
  \item Random Forest, 
  \item Support Vector Machines (SVM), and 
  \item Partial Least Squares-Discriminant Analysis (PLS-DA).
\end{itemize}

Decision trees provide a direct and understandable method for categorization by dividing the feature space using uncomplicated decision rules. Logistic regression is particularly suitable for binary classification tasks and offers probabilistic outputs, which are useful for evaluating the probability of sample infection. Partial least squares discriminant analysis (PLSDA) is highly efficient in managing datasets with a large number of dimensions and has been extensively applied in GC-IMS data analysis. The random forest algorithm is a flexible ensemble technique that utilizes the strength of numerous decision trees to enhance the accuracy of classification and reduce the problem of overfitting. Support vector machines (SVM) are renowned for their capacity to create optimal hyperplanes in spaces with a high number of dimensions. This makes them highly suitable for intricate classification tasks that involve non-linear decision boundaries.

We performed a series of tests to assess the efficacy of each machine learning algorithm in categorizing the GC-IMS data for infection. These studies were conducted to evaluate the potential accuracy of each algorithm in differentiating between infected and non-infected samples and thus to give us a direction on which algorithms are more promising to evaluate more in depth. The following list summaries the experimental setup for each algorithm and Table \ref{tab:algorithms_accuracies} demonstrates the effectiveness in terms of accuracy for each method in managing the intricacy of the high-dimensional GC-IMS data and its capacity for precise infection discovery.

The experimental setup for the evaluation of each algorithm using k-fold cross validation is explained in the following:

\begin{enumerate}
\item Decision Trees:
We trained decision tree classifiers using various hyper-parameters such as maximum depth and minimum samples per leaf, and using feature selection and feature importance
(select top 100 features) to optimize performance.
\item Logistic Regression: Logistic regression models were trained using different regularization techniques (e.g., L1 and L2 regularization) to assess their impact on classification performance. (Selected top 84 features)
\item Random Forest: Random forest classifiers were trained using an ensemble of decision trees with varying tree depths and number of estimators, choosing most important features (top 100).
\item Support Vector Machines (SVM): SVM models were trained using feature selection with LinearSVC and different kernel functions (e.g., linear, polynomial) to capture non-linear relationships in the data.
\item Partial Least Squares-Discriminant Analysis (PLS-DA): PLSDA models were trained with different numbers of components to capture the underlying variation in the GC-IMS data.
\end{enumerate}

Table \ref{tab:algorithms_accuracies} depicts the accuracy scores attained by each machine learning method to give a clear comparison of how well it performed in classifying the GC-IMS data for infection detection.

\begin{table}[h!]
\centering
\caption{Accuracy Scores of Machine Learning Algorithms for GC-IMS Data Classification}
\label{tab:algorithms_accuracies}
\begin{tabular}{|cc|}
\hline
\multicolumn{2}{|c|}{\textbf{Classification Prediction Accuracy}}                     \\ \hline
\multicolumn{1}{|c|}{\textbf{Algorithm}}                          & \textbf{Accuracy} \\ \hline
\multicolumn{1}{|c|}{Decision Trees}                              & 0.61                 \\ \hline
\multicolumn{1}{|c|}{Logistic Regression}                         & 0.53                 \\ \hline
\multicolumn{1}{|c|}{Random Forest}                               & 0.84                 \\ \hline
\multicolumn{1}{|c|}{Support Vector Machines}                     & 0.83                 \\ \hline
\multicolumn{1}{|c|}{Partial Least Squares-Discriminant Analysis} & 0.80               \\ \hline
\end{tabular}
\end{table}

Based on our initial results, depicted in Table \ref{tab:algorithms_accuracies}, it is clear that Singular Value Decomposition (SVD), Random Forest, and Partial Least Squares Discriminant Analysis (PLS-DA) demonstrated the highest level of accuracy among the five machine learning algorithms that were assessed. The algorithms exhibited solid accuracy rates in categorizing the GC-IMS data, highlighting their potential for precise sample classification and diagnosis of infection status. Nevertheless, it is crucial to recognize that these findings are derived from initial testing and constitute an initial evaluation of the algorithm's performance, however, comparing these results with other related works in the literature (e.g. \cite{phillips2014short, tiele2020breath}) that also achieve an accuracy of around 80\%, in different tasks of course, we can identify the need for further exploration on extending our approach to more sophisticated methods and algorithms.

\section{Discussion}

\subsection{Challenges and Considerations}

Although our preliminary tests have demonstrated encouraging outcomes in employing machine learning models for infection detection, there are various issues that must be resolved. An essential obstacle lies in the continuous refinement and enhancement of machine learning models. While we have made progress in achieving high accuracy rates, there is still ample room for improving the performance of these models using deep learning models. The optimization process may require adjusting parameters, choosing appropriate features, and experimenting with larger volumes of data, various algorithms or ensemble techniques to enhance the accuracy of classification and its ability to be applied to new data.

An additional crucial factor to consider is the comprehensive validation and verification of the created models. Conducting thorough validation experiments using multiple datasets that accurately represent real-world clinical settings is crucial. This validation technique guarantees that the machine learning models demonstrate resilience, scalability, and dependability when used with diverse datasets and under differing circumstances. Thorough validation and verification are crucial to determine the effectiveness and usability of the models in clinical practice, ultimately leading to enhanced diagnostic accuracy and patient care outcomes.

\subsection{Future Directions}

The initial achievements in employing advanced machine learning techniques to identify infections create opportunities for promising future developments and progress in this domain. For our future directions, we prioritize improving and refining machine learning models. This includes continuous tasks such as adjusting parameters, selecting features, and investigating new algorithms or ensemble techniques to enhance the precision and resilience of infection detection and classification.

In addition, further research will focus on the advancement of explainability methodologies for machine learning models\cite{rodis2023current}. Gaining insights into infection identification requires a thorough understanding of the components and attributes that drive the decision-making process of the model. Investigating explainability techniques and employing methodologies like those reported in our previous work \cite{ACMComputingSurveys_rodis2023multimodal} will improve the comprehensibility of machine learning models, promoting trust and assurance in their clinical uses.

Ultimately, utilizing advanced analytics techniques like deep learning, will make it possible to analyze spectra images and identify intricate patterns and correlations \cite{bai2021explainable, chen2019looks, papadopoulos2007combining}. This will result in improved diagnostic accuracy and the development of individualized treatment methods.

\section*{Acknowledgment}

The research leading to the results of this paper has received funding from the European Union’s Horizon Europe research and development program under grant agreement No 101073924 (ONELAB).






\bibliographystyle{./IEEEtran}
\bibliography{./main}


\end{document}